\pdfoutput=1

\documentclass[11pt]{article}

\usepackage{ACL2023}
\usepackage{tabularx}

\usepackage{times}
\usepackage{latexsym}
\usepackage{graphicx}
\usepackage[T1]{fontenc}
\usepackage{enumitem}

\usepackage[utf8]{inputenc}

\usepackage{microtype}

\usepackage{inconsolata}

\title{
Reconsidering SMT Over NMT for \textit{Closely Related} Languages: A Case Study of Persian-Hindi Pair}
\author{
Waisullah Yousofi \textnormal{and}  Pushpak Bhattacharyya \\
Computation for Indian Langauge Technology (CFILT)\\
Indian Institute of Technology Bombay, Mumbai, India.\\ 
\texttt{(waisullah, pb)@cse.iitb.ac.in} 
}

\begin{document}
\maketitle
\begin{abstract}

This paper demonstrates that Phrase-Based Statistical Machine Translation (PBSMT) can outperform Transformer-based Neural Machine Translation (NMT) in moderate-resource scenarios, specifically for structurally similar languages, Persian-Hindi pair in our case. Despite the Transformer architecture's typical preference for large parallel corpora, our results show that PBSMT achieves a BLEU score of \textbf{66.32}, significantly exceeding the Transformer-NMT score of 53.7 ingesting the same dataset.

Additionally, we explore variations of the SMT architecture, including training on Romanized text and modifying the word order of Persian sentences to match the left-to-right (LTR) structure of Hindi. Our findings highlight the importance of choosing the right architecture based on language pair characteristics, advocating for SMT as a high-performing alternative in such cases, even in contexts commonly dominated by NMT.
\end{abstract}

\section{Introduction}

In the current state of NLP affairs, the performance of attention-based \cite{bahdanau2014neural, vaswani2017attention} MT systems reaches BLEU scores of almost one. However, the underlying Neural Network (NN) architectures of such high-performing models, assume that the language pairs have a humongous diverse parallel corpora to achieve such desired performance. Of course, there are certain high-source language pairs such as English and French which benefit from those solutions but the MT system of other natural languages that utilize NN architecture without meeting the architecture's assumptions are on the disadvantage side and will generate translations that are way far being accepted by native speakers.

Beyond the need for large datasets, another great concern when using NNs is their high power consumption and the environmental impact they leave behind—through processes such as training, inference, and experimentation, which all contribute to carbon footprints \cite{faiz2023llmcarbon}.

To walk through an efficient alternative path, we looked at the big picture of natural languages, focusing on their linguistic families and the factors that group languages. We have observed that the property of linguistic closeness of less divergent languages can be exploited. The key contributions of our paper are:
\begin{itemize}[nosep]
    \item The first attempt to build a general domain MT system of Persian-Hindi languages. 
    \item We demonstrate that for structurally close language pairs having a moderate-sized (1M+ sentences) high-quality parallel corpus, SMT outperforms a Transformer-based NMT model.    
    \item Suggesting alternative paths to build computationally and environmentally efficient MT systems.
\end{itemize}

The upcoming sections are structured as follows: section 2 presents a review of the literature, followed by a detailed description of the parallel corpus used in our experiments and analysis in section 3. Section 4 outlines the experimental setup, while section 5 provides a comprehensive analysis of the results. Finally, section 6 concludes the paper and discusses potential directions for future research.

\section{Related Works}
Before the revolutionization of the field by the Transformer architecture \cite{vaswani2017attention}, the notion of language closeness was being leveraged in various forms for different language pairs. In this section, we will see that our work is not only different from the perspective of studying a new language pair, Persian-Hindi, which to date no formal research has been conducted yet for the pair, but it also varies in terms of past Transformer comparison of the two architectures, NMT and SMT, given that we have access to a moderate amount of parallel sentences.

\begin{table}[htbp]
\centering
\begin{tabularx}{\linewidth}{|X|X|X|X|}
\hline
\textbf{Split} & \textbf{Sentences} & \textbf{FA-Tokens} & \textbf{HI-Tokens} \\
\hline
Train & 1,01M & 17.25M & 17.75M \\
\hline
Test & 3,000 & 50k & 52K \\
\hline
Tune & 8,000 & 1.36M & 1.39M \\
\hline
\textbf{Total} & \textbf{1M+} & \textbf{19.1M} & \textbf{19.1M} \\
\hline
\end{tabularx}
\caption{Corpus statistics after applying LABSE filtration with a threshold of 0.9.}
\label{tab:filtered_corpora}
\end{table}

\begin{table}[htbp]
\centering
\begin{tabularx}{\linewidth}{|X|X|X|}
\hline
\textbf{Language} & \textbf{Sentences} & \textbf{Tokens}\\
\hline
Persian & 13.7M+ & 190M+\\
\hline
Hindi & 13.7M+ & 207M+ \\
\hline
\end{tabularx}
\caption{Details of normalized but unfiltered monolingual of Persian and Hindi.}
\label{lm-table}
\end{table}

Previous works such as,\cite{toral-way-2015-translating},  hypothesize that translations between related languages tend to be more literal, with complex phenomena (e.g., metaphors) often transferring directly to the target language. In contrast, these phenomena are more likely to require complex translations between unrelated languages. Other instances such as \cite{rios-sharoff-2015-obtaining}, \cite{kunchukuttan-etal-2017-utilizing}, and \cite{kunchukuttan-bhattacharyya-2017-learning}
utilize lexical similarities. Except \cite{jauregi-unanue-etal-2018-english} which shows that for the scenario of low-resource (unlike our scenario which assumes medium resource) languages, SMT performs better than NMT, there is no comparative analysis of the NMT and SMT architectures for structurally similar languages with an assumption concerning the size of the parallel corpus.

\section{Dataset and Preprocessing}

\begin{table*}[htbp]
\centering
\begin{tabular}{|l|l|l|l|}
\hline
\textbf{Corpus} & \textbf{Sentences} & \textbf{FA Tokens} & \textbf{HI Tokens} \\ \hline
CCMatrix v1 & 2,7M+ & 30M+ & 32M+ \\ \hline
NLLB v1 & 2M+ & 30M+ & 32M+ \\ \hline
MultiCCAligned v1.1 & 1M+ & 22M+ & 21M+ \\ \hline
XLEnt v1.2 & 0.4M+ & 1M+ & 1M+ \\ \hline
Tanzil v1 & .01M+ & 4M+ & 4M+ \\ \hline
KDE4 v2 & 72K & 0.3M+ & 0.3M+ \\ \hline
OpenSubtitles v2018 & 48K & 0.2M+ & 0.3M+ \\ \hline
TED2020 v1 & 41K & 0.7M+ & 0.7M+ \\ \hline
GNOME v1 & 40K & 0.1M+ & 0.1M+ \\ \hline
WikiMatrix v1 & 20K+ & 0.3M+ & 0.3M+ \\ \hline
NeuLab-TedTalks v1 & 16K & 0.3M+ & 0.3M+ \\ \hline
QED v2.0a & 2k & 0.7M & 0.6M \\ \hline
ELRC-wikipedia\_health v1 & 1k & 1K & 1K \\ \hline
GlobalVoices v2018q4 & 139 & 1K & 1K \\ \hline
TLDR-pages v2023-08-29 & 58 & 447 & 454 \\ \hline
Wikimedia v20230407 & 40 & 2k & 2K \\ \hline
Ubuntu v14.10 & 6k+ & 27K & 29K \\ \hline
LSCP Corpus   & 4.6M+ & 1.3M+ & 1.1M+ \\ \hline

\textbf{Total} & \textbf{10.9M+} & \textbf{90.9M+} & \textbf{93.7M+} \\ \hline
\end{tabular}

\caption{\label{corpora-numbers}
Basic Statistics of Individual and Merged Corpora.}
\end{table*}

In addition to the "Large Scale Colloquial Persian Dataset" (LSCP) \cite{abdi-khojasteh-etal-2020-lscp}, the other datasets we utilized are primarily sourced from OPUS \cite{tiedemann2016opus}, a well-known repository for parallel corpora of various domains for a vast number of language pairs.

Table \ref{corpora-numbers} shows the basic statistics related to all the corpora. After downloading those \textbf{10.9M+} sentences, we noticed that most of them were of low quality. To filter them we used LABSE \cite{feng2020language} during which \cite{batheja-bhattacharyya-2022-improving}'s work was of help. Before the LABSE-filtration, the preprocessing steps for each corpus include, the removal of empty lines, punctuations, emojis, and deduplication of repeated parallel sentence pairs, normalization, and tokenization using language-specific libraries, 
\verb|indic-nlp-library| 
\cite{indicnlp} for Hindi and \verb|ParsiNorm| \cite{oji2021parsinorm} for Persian.

    An additional time-taking step applied to the LCSP corpus was to pair the sentences first and then pass to the preprocessing phase. Then, we performed LABSE filtration which the dramatic reduction of the original corpus is shown in Table \ref{tab:filtered_corpora}. In one of our SMT experiments, as we will see the details in the next section, the parallel sentences need to be Romanized, for which we employed \verb|uroman| library \cite{hermjakob-etal-2018-box}. For all NMT experiments, the first step after receiving the raw data (Table \ref{tab:filtered_corpora}) was to apply Byte Pair Encoding (BPE) \cite{sennrich-etal-2016-neural} with 32K merge operations.

\begin{figure}[htbp]
  \centering
  \includegraphics[width=1\columnwidth]{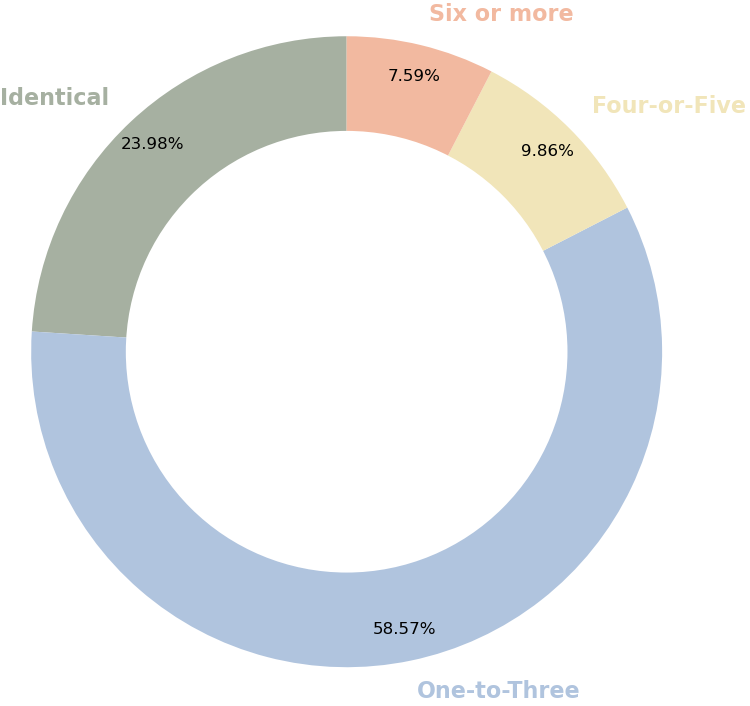}  
  \caption{Categories the differences of lengths of parallel sentences length counted in terms of tokens.
  }
  \label{fig:donut-chart}
\end{figure}

It should be mentioned that for the Language Model (LM) component of the SMT model, we used the unfiltered monolingual of the target language, Hindi, and the corresponding numbers are detailed in Table \ref{lm-table}.

To evaluate the structural similarity between Persian and Hindi, we analyzed the differences in sentence lengths (measured in token counts) across all parallel sentences. We assumed that if the majority of sentences exhibit a difference of three tokens or less, the alignment achieved through \verb|mgiza| would represent an optimal one-to-one correspondence. As illustrated in the donut chart in Figure \ref{fig:donut-chart}, more than 72\% of the parallel sentences in our dataset, Table \ref{tab:filtered_corpora}, have a length difference of less than three tokens. Moreover, through the \textit{\textbf{language divergence} (the phenomenon of languages expressing meaning in divergent ways)} setting proposed by \cite{dorr1993machine}, we studied the fact that the Persian-Hindi pair is almost isomorphic.

From the perspective of structure and syntax of German, Spanish, and English, Dorr proposes a set of seven types of divergences \cite{bhattacharyya2015machine}. For our pair, we examine some types of syntactic divergence through examples provided in the Appendix~\ref{sec:appendix}:
\section{Experimental Setup}
\label{sec:setup}
\subsection{Moses SMT}

\begin{figure}[htbp]
  \centering
  \includegraphics[width=\columnwidth]{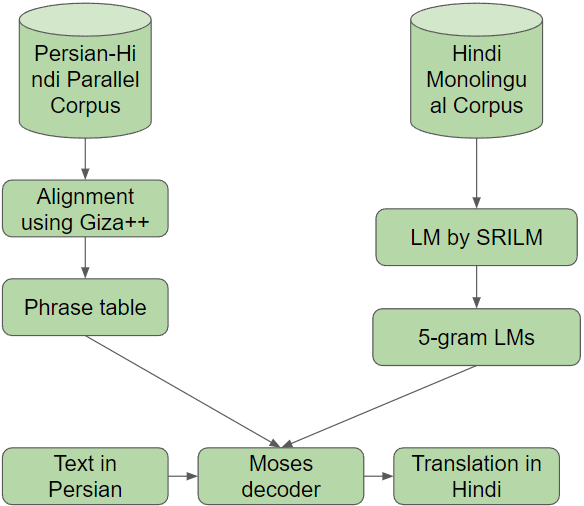}
  \caption{Architecture of Persian to Hindi SMT Model.}
  \label{fig:smt_arch}
\end{figure}

SMT which gave rise to NMT has been around for quite a long time. The basic idea of SMT is to learn the word alignment first and then expand it to phrases to build a phrase table that will be used for predictions. All three SMT-based experiments that we performed generally follow the same pipeline which is illustrated in Figure \ref{fig:smt_arch}. We utilized the open-source toolkit, Moses \cite{koehn2007moses} to train a PBSMT model. First, a word alignment model between the Persian and Hindi languages was trained on the training data using MGIZA++ toolkit \cite{och2000improved}. Next, a 5-gram Language Model (LM) employing Kneser-Ney smoothing and interpolation was built using \textit{SRILM} toolkit developed by \cite{kneser1995improved}. Since these two languages do not require transforming their scripts into lower-case, true-case, etc, we neither applied those transformations nor used Moses's default tokenizer- we used language-specific libraries for better results. Finally, Moses decoder was used to translate sentences based on these components.

\subsection{OpenNMT}
We fed the same data splits, Table \ref{tab:filtered_corpora}, that were used for SMT, to an NMT Transformer \cite{vaswani2017attention} model with the help of open-source OpenNMT \cite{klein2017opennmt} library. The NMT model consists of 8 layers of encoder and decoder each with 8 attention heads,  the embedding layer of size 512 with positional encoding enabled. Coming to hyperparameters, the batch size was set to 4096 tokens iterating over 300K steps with an initial learning rate of 2 along with Adam optimizer setting $\beta_1$ and $\beta_2$ to 0.9 and 0.998 respectively. Additionally, we utilized 8000 warmup steps.

In terms of GPU usage, two NVIDIA GeForce RTX 2080 GPUs were occupied during training.

\subsection{Evaluation}
Throughout the experiments, we used the BLEU metric \cite{papineni2002bleu}. Although, the way Moses \footnote{https://github.com/moses-smt/mosesdecoder/blob/master/scripts/generic/mteval-v13a.pl} calculates this score is correct 
but a refined version that better handles BLEU's hyperparameters and computes it is \verb|SACREBLEU|  \cite{post-2018-call}.

\section{Experiments and Results}

The first SMT experiment used the normalized filtered data of Table \ref{tab:filtered_corpora}, applying it conventionally to SMT-Moses with the configurations detailed in Section~\ref{sec:setup}. While Moses was processing the data, we simultaneously set up our encoder-decoder Transformer model. The best NMT model achieved \textit{53.7}, whereas the initial SMT model had a BLEU score \textbf{64.9}. To verify the high BLEU score of SMT, we conducted a 4-fold cross-validation the BLEU scores of which are \textit{\textbf{67.32}, 66.32, 64.90,} and \textit{66.74}, respectively. Therefore, our best SMT model's BLEU has been marked \textbf{66.32}- the average of 4-fold's BLEU. Also, by looking at the algorithmic nature of each architecture, it makes sense for the SMT to perform better than NMT in the existence of moderate data size. Because, SMT uses Expectation Maximization (EM) algorithm for alignment, and since the source and target languages are almost always one-to-one mapping, we need less data-size than that of NN. 

Since Persian and Hindi share many common words, in our second experiment, parallel sentences were first Romanized to increase text similarity \cite{hermjakob-etal-2018-box}. However, the BLEU score dropped to \textit{51.21} from \textit{66.7}. One highly probable reason is due to the diacritics (mark that is placed above, below, or through a letter to indicate how it should be pronounced) that some Persian words have in order to determine the phoneme of a word and hence the associated meaning. For example, \verb|gul (flower)| and \verb|gel (mud)| are two words the sound and meaning of which can be determined from the context (without diacritics) or using diacritics. Romanization often results in the loss of these nuances, leading to ambiguities in alignment and translation.

\begin{table}[h!]
\centering
\begin{tabular}{|l|c|}
\hline
\textbf{Model}                          & \textbf{BLEU Score} \\ \hline
Initial SMT Model      & 64.91              \\ \hline
Best SMT Model   & \textbf{66.32}              \\ \hline
FOLD 1                                    & {\textbf{67.32}}       \\ 
FOLD 2                                    & 66.32              \\ 
FOLD 3                                    & 64.90              \\ 
FOLD 4                                    & 66.74              \\ \hline
NMT Transformer Model                           & 53.7               \\ \hline
Romanized-SMT Model                             & 51.21              \\ \hline
Inverted-SMT Model          & 48.74              \\ \hline
\end{tabular}
\caption{BLEU scores of various SMT and NMT models.}
\label{tab:bleu_scores}
\end{table}

In the final experiment, we reversed the Persian scripts from right-to-left (RTL) to left-to-right (LTR) to align the writing direction of Hindi, expecting improved alignment quality. Unfortunately, this resulted in a further decrease in the BLEU score, falling to \textit{48.74}. This decline can be attributed to the fact that reversing a sentence alters its meaning. Although both Persian and Hindi are classified as free-word-order languages, we observed that the phenomenon where “only constructs that follow each other can be moved to any other position in the sentence while still preserving meaning” is compromised when inversion occurs, leading to a change in interpretation. See Appendix~\ref{sec:appendix} for examples. Table \ref{tab:bleu_scores} summarizes the BLEU scores for the different experiments we performed.

\section{Conclusion and Future Works}
This study presents a comparative analysis of SMT and NMT for the Persian-to-Hindi language pair. Our findings demonstrate that SMT yields superior results in closely related languages, attributable to their shared linguistic structures. Additionally, we observed that reversing the order of Persian sentences from RTL to LTR negatively impacted the SMT model's performance, resulting in a loss of meaning. In contrast, romanizing the input text showed a beneficial effect compared to the inversion experiment.

Future work will focus on deepening our understanding of these languages and exploring alternative approaches, such as translation through common space word embedding, transfer learning, and pivot-based NMT with English as a bridging language.

\bibliographystyle{acl_natbib}  
\bibliography{main}  
\appendix
\section{Examples Appendix}
\label{sec:appendix}

\subsection{Paralle Sentences of Different Length Categories}

\subsubsection{Identical Lengths}


\noindent\begin{tabular}{@{}l@{\hspace{0em}}p{0.80\linewidth}@{}}
{1.1.1.Fa:} & \texttt{tamam mahsulat hamel shodeh 100\% bazorsi mi shvand.} \\
{1.1.1.Hi:} & \texttt{bheje gae sabhee utpaad 100 nireekshan kie jaate hain.} \\
{1.1.1.En:} & \texttt{All shipped products will be 100\% inspected.} \\
\end{tabular}

\vspace{0.2cm}
\noindent\begin{tabular}{@{}l@{\hspace{0em}}p{0.80\linewidth}@{}}
{1.1.2.Fa:} & \texttt{baraye etlaat bishtar lotfa ba ma tamas begirid} \\
{1.1.2.Hi:} & \texttt{adhik jaanakaaree ke lie krpaya hamase sampark karen} \\
{1.1.2.En:} & \texttt{For more information please contact us} \\
\end{tabular}
\\

\subsubsection{One-to-Three Token Difference}

\noindent\begin{tabular}{@{}l@{\hspace{0em}}p{0.80\linewidth}@{}}
{1.2.1.Fa:} & \texttt{akharin ghimet sakeh ve tala dar bazar. [7-tokens]}\\
{1.2.1.Hi:} & \texttt{baajaar par naveenatam sikka aur sone kee keematen. [8-tokens]}\\
{1.2.1.En:} & \texttt{The latest price of coins and gold in the market.}\\
\end{tabular}
\vspace{0.2cm}

\noindent\begin{tabular}{@{}l@{\hspace{0em}}p{0.80\linewidth}@{}}
{1.2.2.Fa:} & \texttt{Har zemestān bahāri dar pay dārad. [6-tokens]}\\
{1.2.2.Hi:} & \texttt{Har sardī ke baad vasant ṛtu hotī hai. [8-tokens]}\\
{1.2.2.En:} & \texttt{After every winter there's spring.}\\
\end{tabular}
\\

\subsubsection{Four-or-Five Token Difference}
\noindent\begin{tabular}{@{}l@{\hspace{0em}}p{0.80\linewidth}@{}}
{1.3.1.Fa:} & \texttt{pish bini ab ve npava dar litvania.[7-tokens]}\\
{1.3.1.Hi:} & \texttt{Havāmāna andāja lithu'āniyā. [3-tokens]}\\
{1.3.1.En:} & \texttt{Weather forecast in Lithuania.}\\
\end{tabular}

\vspace{0.2cm}

\noindent\begin{tabular}{@{}l@{\hspace{0em}}p{0.80\linewidth}@{}}
{1.3.2.Fa:} & \texttt{besiar sadeh baraye estefadeh.[4-tokens]}\\
{1.3.2.Hi:} & \texttt{ka upayog karane ke lie bahut hee saral. [8-tokens]}\\
{1.3.2.En:} & \texttt{Very simple to use.}\\
\end{tabular}
\\

\subsection{Inversion Example}

\noindent\begin{tabular}{@{}l@{\hspace{0em}}p{0.80\linewidth}@{}}
{2.1.original-Fa  (read from right-to-left):}\\
\verb|.daram arezo azizan baraye zibayi|\\
{2.1.En (of original-Fa ):}\\
\verb|I wish beauty for the loved ones.|\\
{2.1.inverted-Fa inverted (read left-to-right):}\\
\verb|daram arezo azizan baraye zibayi.|\\
{2.1.En (of inverted-Fa ):} \\
\verb|I wish the loved ones for beauty.|\\
\end{tabular}
\\

As we can see, the inverted sentence wishes \textit{the loved ones} FOR \textit{the beauty}, which to some extent does not make sense at all. Hence, the conclusion we made here is that the inversion of the sentence disrupts the intended meaning and perhaps alignment, which consequently affects the overall performance negatively.

\subsection{Syntactic Divergence}
Through the examples taken from \cite{bhattacharyya2015machine}, we are going to observe some cases where Persian and Hindi sentences do not diverge, which implies syntactic closeness.

\subsubsection{Constituent Order Divergence}
It is related to the divergence of word order between a pair. For instance, below we can see that both follow the same SOV order.
\noindent\begin{tabular}{@{}l@{\hspace{0em}}p{0.80\linewidth}@{}}
{3.1.1.En:} & \texttt{Jim (S) is playing (V) tennis (O)}\\
{3.1.1.Fa:} & \texttt{jim (S) tenis (O) bazi karde rahi ast(V). [Jim tenis play done being is]}\\
{3.1.1.Hi:} & \texttt{jeem (S) tenis (O) khel rahaa hai (V) [Jim tennis playing is]}\\
\end{tabular}
\\

\subsubsection{Null Subject Divergence}
Null Subject Divergence refers to the phenomenon languages, such as Persian and Hindi, omit the subject pronoun (like "there" in English), because the subject is implied or understood from the verb form or context.

\noindent\begin{tabular}{@{}l@{\hspace{0em}}p{0.80\linewidth}@{}}
{3.2.1.En:} & \texttt{Long ago, there was a king}\\
{3.2.1.Fa:} & \texttt{Khili vaqt pish, yek padshah bud.[Long ago one king was]}\\
{3.2.1.Hi:} & \texttt{bahut pahale ek raajaa thaa [Long ago one king was]}\\
\end{tabular}
Similar practices can be performed to observe that both languages diverge only rarely- conflational divergence.
\end{document}